\documentclass[letterpaper, 10 pt, conference]{ieeeconf}  

\IEEEoverridecommandlockouts                              

\usepackage{graphics} 
\usepackage{epsfig} 
\usepackage{times} 
\usepackage{amsmath} 
\usepackage{amssymb}  
\usepackage[backend=biber,sorting=nyt,sortcites=true]{biblatex}
\usepackage{multirow}
\usepackage{multicol}
\usepackage{hyperref}
\usepackage{booktabs}
\usepackage{enumerate}
\usepackage{color}
\usepackage{tabularx}
\usepackage{hyperref}
\usepackage{pifont}
\usepackage{algorithm}
\usepackage{algpseudocode}
\usepackage{amsmath}
\usepackage{adjustbox}
\usepackage[usenames,dvipsnames,table]{xcolor}
\hypersetup{
    colorlinks=true,
    linkcolor=MidnightBlue,
    filecolor=magenta,      
    urlcolor=MidnightBlue,
    citecolor=MidnightBlue,
} 
\usepackage[font=small,labelfont=bf]{caption}

\definecolor{darkred}{rgb}{0.76, 0.23, 0.13}
\definecolor{darkgreen}{rgb}{0.01, 0.75, 0.24}
\definecolor{darkgray}{rgb}{0.66, 0.66, 0.66}

\title{\LARGE \bf
Learning Dexterous Manipulation with Quantized Hand State
}

\author{
Ying Feng$^{*, 1, 3}$, Hongjie Fang$^{*, 1, \dagger}$, Yinong He$^{*, 1, 4}$,\\ Jingjing Chen$^1$, Chenxi Wang$^2$, Zihao He$^1$, Ruonan Liu$^1$, Cewu Lu$^{1,2,3,\dagger}$ %
\thanks{$^*$Equal Contribution.\quad $^\dagger$Corresponding Authors.}
\thanks{$^1$Shanghai Jiao Tong University. \quad $^2$Noematrix.}
\thanks{$^3$Shanghai Innovation Institute. \quad $^4$Carnegie Mellon University.}
}

\begin{document}

\maketitle
\thispagestyle{empty}
\pagestyle{empty}

\begin{abstract}

Dexterous robotic hands enable robots to perform complex manipulations that require fine-grained control and adaptability. Achieving such manipulation is challenging because the high degrees of freedom tightly couple hand and arm motions, making learning and control difficult. Successful dexterous manipulation relies not only on precise hand motions, but also on accurate spatial positioning of the arm and coordinated arm-hand dynamics. However, most existing visuomotor policies represent arm and hand actions in a single combined space, which often causes high-dimensional hand actions to dominate the coupled action space and compromise arm control. To address this, we propose \textit{DQ-RISE}, which quantizes hand states to simplify hand motion prediction while preserving essential patterns, and applies a continuous relaxation that allows arm actions to diffuse jointly with these compact hand states. This design enables the policy to learn arm-hand coordination from data while preventing hand actions from overwhelming the action space. Experiments show that \textit{DQ-RISE} achieves more balanced and efficient learning, paving the way toward structured and generalizable dexterous manipulation. Project website: \href{http://rise-policy.github.io/DQ-RISE/}{http://rise-policy.github.io/DQ-RISE/}.

\end{abstract}

\section{Introduction}\label{sec:intro}

Dexterous robotic hands have become a central focus in robotics, allowing robots to interact with the physical world in ways that resemble human manipulation~\cite{dex-survey, eyesight_hand, leap_hand}. With multiple independently actuated fingers and rich contact surfaces, they can perform complex manipulations beyond parallel grasping~\cite{anygrasp, graspness}, including precise in-hand reorientation~\cite{dex-inhand, inhand-re, visual-dex} and adaptive gripping for objects of diverse shapes and sizes~\cite{dexgraspnet, anydexgrasp, unidexgrasp}. Beyond these fundamental tasks, dexterous hands can use tools, coordinate bimanual actions, and carry out intricate assembly procedures~\cite{bidexhands, dex-assembly, dex-tool}, bridging the gap between rigid robotic actuation and human-like flexibility. Together, these capabilities define the scope and promise of robotic dexterity.

\begin{figure}[t]
    \centering
    \includegraphics[width=\linewidth]{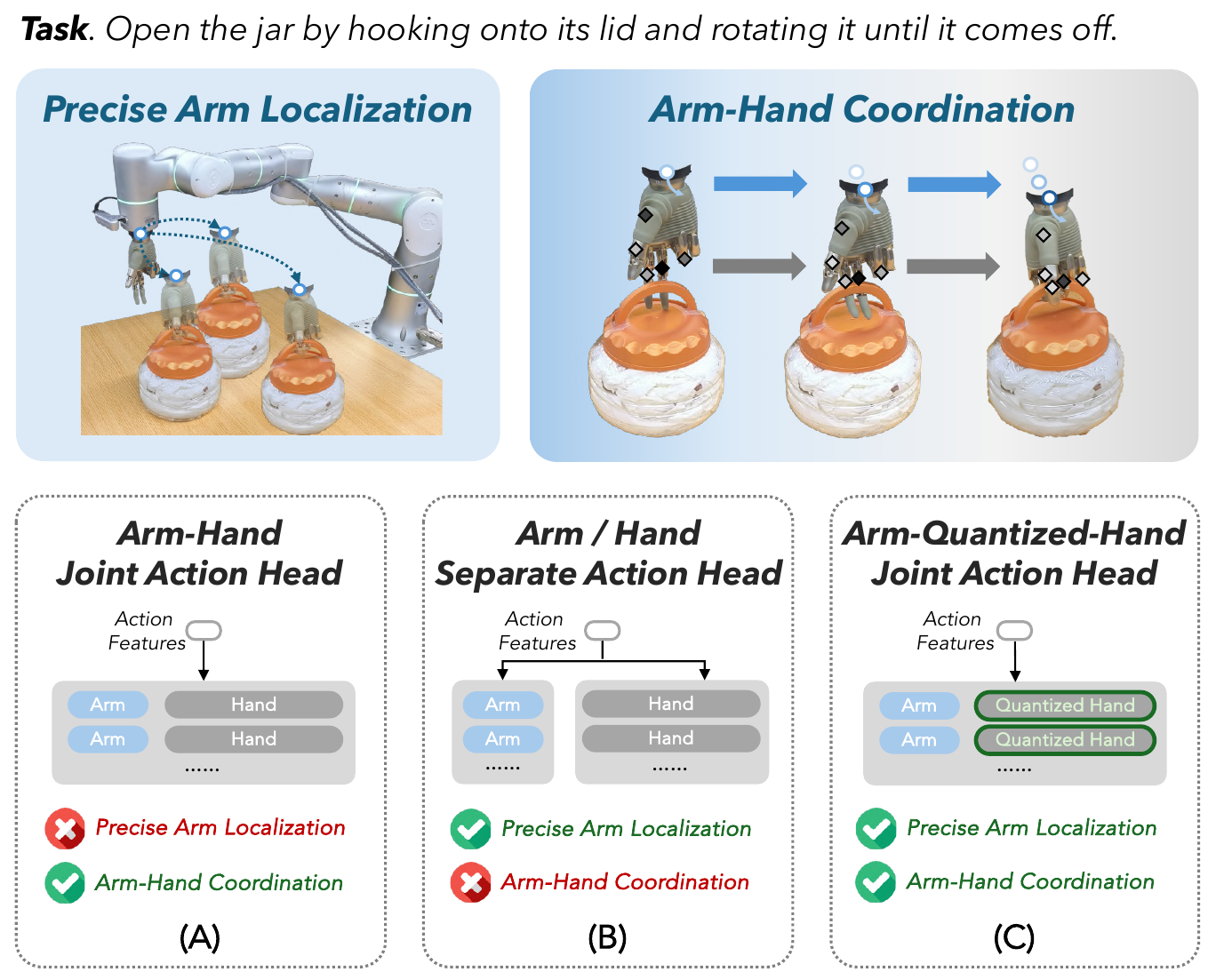}
    \caption{\textbf{Dexterous Manipulation from the Action Prediction Perspective.} Beyond hand motion, successful dexterous manipulation also requires precise arm localization and coordinated arm-hand dynamics. \textbf{(A)} Existing visuomotor policies predict arm and hand actions jointly, causing hand actions to dominate the combined action space and arm localization to suffer. \textbf{(B)} Naively separating arm and hand predictions can lead to incoherent coordination. \textbf{(C)} Our approach quantizes hand states to preserve hand motion while jointly diffusing arm actions, enabling precise arm localization and smooth arm-hand coordination.}
    \label{fig:teaser}
    \vspace{-0.6cm}
\end{figure}

However, these abilities come at a cost: the additional degrees of freedom (DoF) make manipulation highly complex and tightly coupled with arm motion. To illustrate, consider the everyday task of opening a jar by hooking onto its lid and rotating it until it comes off, as depicted in Fig.~\ref{fig:teaser}. Successful manipulation requires not only \textit{accurate hand motions} to twist the lid, but also two often overlooked aspects regarding the arm: (1) \textit{precise arm localization} ---bringing the end-effector exactly to the lid’s position; and (2) \textit{arm-hand coordination} --- when the fingers hook the lid, the arm must simultaneously press downward to prevent lifting the jar. Together, these requirements highlight that beyond finger dexterity, the primary goals of dexterous manipulation include spatial accuracy of the arm and cooperative dynamics between arm and hand.

From the perspective of robot action prediction, visuomotor policies should output both arm and hand actions. Most existing policies treat them together in a single combined action space~\cite{hato, dexcap, bidex, vitacformer, dexos}, as shown in Fig.~\ref{fig:teaser}A. While convenient, this approach often leads to imbalanced learning: the high-DoF hand actions dominate the combined action space and hinder accurate arm control, as confirmed by our experiments in \S\ref{sec:exp-result}. This observation motivates a clearer functional distinction: \textit{robotic arms primarily handle spatial localization, while dexterous hands are responsible for executing fine-grained actions}. Accordingly, visuomotor policies should \textit{focus on spatial localization for the arm} while \textit{memorizing action patterns for the dexterous hand}. However, naively disentangling arm and hand action generation, as illustrated in Fig.~\ref{fig:teaser}B, can break their coordination and limit overall policy performance.

Building on this insight, we propose \textit{\textbf{DQ-RISE}}, an extension of the base visuomotor policy RISE~\cite{rise} that introduces structured action prediction for dexterous manipulation. We quantize dexterous hand states into a compact set of task-relevant patterns, reformulating hand action prediction as a classification problem analogous to gripper open/close control. To maintain arm-hand coordination, we further introduce a continuous relaxation process that enables the policy to diffuse arm actions jointly with quantized hand actions~\cite{ddpm}, thereby preserving fine-grained dexterity while improving arm localization. In addition, we design a hybrid dexterous teleoperation system for demonstration collection, which facilitates intuitive arm-hand control compared to existing systems. Extensive experiments across diverse dexterous tasks demonstrate that \textit{\textbf{DQ-RISE}} achieves more balanced and efficient learning than other action prediction schemes, highlighting a practical pathway toward scalable and generalizable robotic dexterity.

\section{Related Works}

\subsection{Dexterous Manipulation}

Dexterous manipulation has recently attracted considerable attention in robotics. Owing to the high DoF of dexterous hands, prior works have primarily relied on reinforcement learning to acquire complex manipulation skills~\cite{lin2025sim, CaggianoDK23, Zhu0RLK19, bidexhands, QinHY0022, dex-inhand, inhand-re, visual-dex}. These approaches typically learn state-based policies in simulation with carefully designed reward functions~\cite{NagabandiKLK19}. To bridge the gap between simulation and the real world, where observations differ and privileged state information is unavailable, researchers have applied techniques like sim-to-real transfer~\cite{lin2025sim, QinHY0022} and teacher-student distillation~\cite{visual-dex, clutterdexgrasp} to improve the policy's adaptability.

Another line of work learns dexterous hand behaviors from human demonstrations~\cite{dexmv, maniptrans} or teleoperated demonstrations~\cite{hato,dexcap, bidex, vitacformer, dexos} within the imitation learning framework. However, as discussed in \S\ref{sec:intro}, existing imitation policies mostly predict arm and hand actions jointly in a naively-combined high-DOF action space, often resulting in unbalanced learning and suboptimal hand-arm coordination. To address this limitation, we propose disentangling the joint prediction problem into a more principled and tractable formulation by leveraging the distinct properties of arm motions and hand actions.

\subsection{Quantization in Robotics}

Quantization in robotics can be broadly categorized into two main directions: observation quantization and action discretization. Common tools include VQ-VAE~\cite{vqvae,vqvae2,resvqvae} and VQ-GAN~\cite{vqgan}, which leverage generative models~\cite{vae,gan} and discrete codebooks to produce discrete representations of high-dimensional data, enabling more efficient learning.

Observation quantization compresses high-dimensional sensory inputs like images into discrete latent codes. Visuomotor policies and vision-language-action models often leverage this for future observation prediction~\cite{gr2, moto, upvla, lapa, molmoact, uva}, as predicting latent codes of observations is more tractable and useful than forecasting their raw pixel values~\cite{uva}. This future prediction serves as an auxiliary task that supports action generation and reasoning~\cite{molmoact}. Action quantization discretizes continuous control signals into representative primitives or codebook entries~\cite{vqvla, vqbet, vqace}. This simplification improves learning stability and overall performance. However, most existing approaches focus on action chunks that model future robot motion~\cite{act}.

In contrast, we propose to quantize the dexterous hand state (single-step action) into compact and meaningful discrete codes. Rather than modeling motion trajectories, we directly model hand states, making the quantization more explainable and intuitive. Empirically, we also found that quantizing action chunks often produces less meaningful codes and worse performance compared to state quantization.

\subsection{Dexterous Teleoperation System}

Compared with teleoperation systems for grippers~\cite{airexo, rise2, rh20t, act, umi, gello}, dexterous hand teleoperation is substantially more challenging because operators must control high-DoF configurations, coordinate many coupled joints, and handle complex contact dynamics while also managing arm motions. Existing approaches tackle this in different ways: some introduce hardware that directly maps human hand motion to robotic joints~\cite{doglove, dexumi, tilde, dexos, bidex}, while others employ hand pose estimation~\cite{open-television, dart, kamijo2024learning, ace, anyteleop} or motion-capture gloves~\cite{dexcap} to capture human keypoints, and then retarget them to the dexterous hand~\cite{geort, dexmachina}. For joint hand-arm teleoperation, most systems rely on VR/AR joystick-based interfaces for arm control~\cite{open-television, dart, kamijo2024learning, typetele, hato, dexcap}, while a few utilize exoskeletons for full-arm mapping~\cite{dexos, bidex, ace}.

Recently, several studies have simplified human control by defining discrete gestures during data collection~\cite{typetele, hato}. While this is conceptually similar to our insight of quantizing hand states, the key difference is that their approach relies on \textit{manual quantization at collection time}, whereas we apply \textit{automatic quantization during policy training}. Enforcing discrete gestures during collection can increase the cognitive load on operators, bias their control strategies, and limit the advantages of high-DoF dexterous hands over low-DoF end-effectors, often leading to unnatural demonstrations. To this end, we let operators freely control the hand during data collection and perform quantization afterwards for effective and interpretable policy learning.

\section{Method}

\begin{figure}[t]
    \centering
    \includegraphics[width=0.95\linewidth]{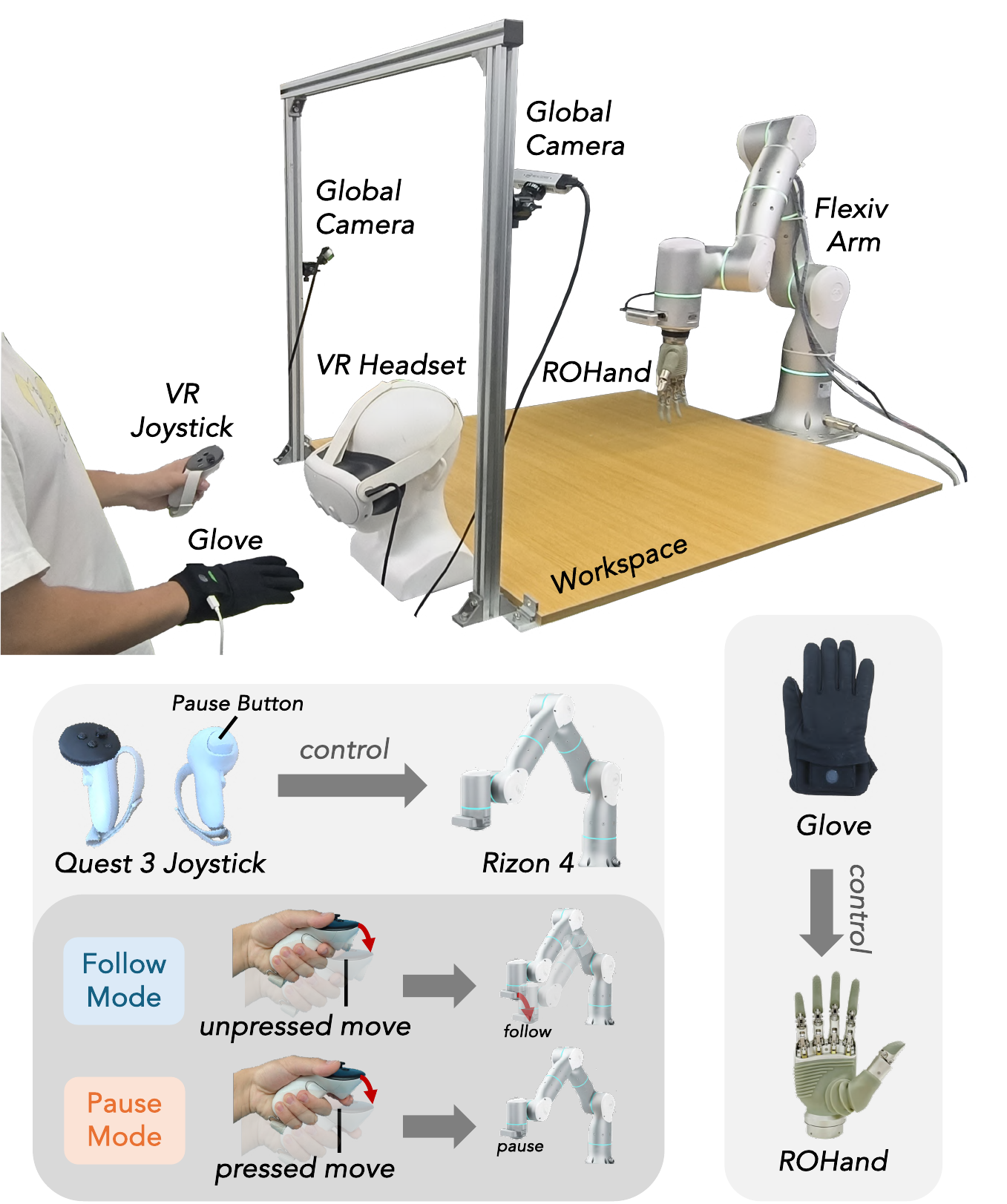}
    \caption{\textbf{Robot Platform and Hybrid Dexterous Teleoperation System.} Our platform consists of a Flexiv robotic arm equipped with an ROHand. During teleoperation, the arm is controlled via a VR joystick, where the joystick button can be used to pause arm motion and adjust the joystick pose for more intuitive and convenient operation. For hand control, we use a GForce glove to directly operate the ROHand using joint correspondence.}
    \label{fig:system}
    \vspace{-0.3cm}
\end{figure}

\begin{figure*}
    \centering
    \includegraphics[width=\linewidth]{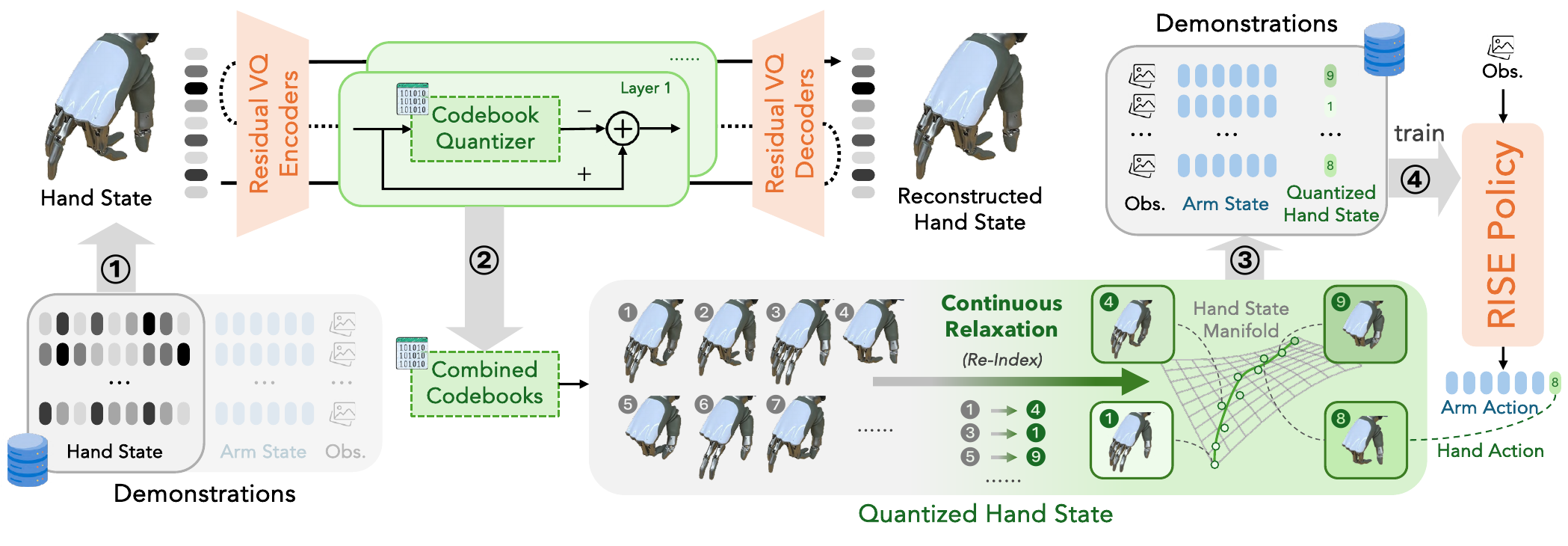}
    \caption{\textbf{\textit{DQ-RISE} Policy Architecture.} \textcircled{\raisebox{-0.9pt}{1}} Hand state data from demonstrations are used to train a residual VQ-VAE~\cite{resvqvae} for hand state quantization (\S\ref{sec:method-quantize}); \textcircled{\raisebox{-0.9pt}{2}} The trained codebooks yield $K$ quantized hand states, which are re-indexed to maintain consistency between consecutive codes and sequential continuity across all codes (\S\ref{sec:method-continuous}); \textcircled{\raisebox{-0.9pt}{3}} The original hand states/actions are replaced by these re-indexed states in the demonstration dataset (\S\ref{sec:method-policy});  \textcircled{\raisebox{-0.9pt}{4}} The visuomotor policy is trained on the transformed dataset, jointly diffusing arm and hand actions; during inference, the predicted continuous hand actions are projected to the nearest quantized actions for execution (\S\ref{sec:method-policy}).}
    \label{fig:policy}
    \vspace{-0.4cm}
\end{figure*}

In this section, we first tackle data collection by presenting our VR-glove hybrid dexterous teleoperation system (\S\ref{sec:data-collection}). We then quantize hand states into discrete latent codes (\S\ref{sec:method-quantize}). While a straightforward approach is to treat these discrete states directly as a classification problem for policy learning, we find that jointly diffusing arm and hand actions improves action consistency and overall performance. Hence, we re-index the discrete hand states for continuous relaxation (\S\ref{sec:method-continuous}) and use the resulting relabeled demonstrations to train the base visuomotor policy, ensuring more accurate and robust hand-arm coordination during manipulation (\S\ref{sec:method-policy}). An overview of our policy is illustrated in Fig.~\ref{fig:policy}.

\subsection{Hybrid Dexterous Teleoperation System}\label{sec:data-collection}

As shown in Fig.~\ref{fig:system}, our platform consists of a Flexiv Rizon 4 robotic arm with a 6-DoF OyMotion ROHand.
Two Intel RealSense D415 cameras provide global observations to mitigate occlusions, while a wrist-mounted Intel RealSense D435 camera is used for calibration only. 

We design a VR-glove hybrid system for teleoperation with dexterous hands in a single-arm setup. A Meta Quest 3 VR headset tracks its joystick pose to control the arm movements, while the operator uses the other hand with a glove to capture precise hand motions for dexterous hand manipulation, as shown in Fig.~\ref{fig:system}. For the ROHand, we use an OyMotion GForce glove, which measures the human hand motions corresponding to each dexterous hand joint using pressure tablet sensors, and maps these measurements into joint signals to control the dexterous hand. 

Compared to pure vision-based systems~\cite{open-television, ace, dart, kamijo2024learning} and discretized gesture systems~\cite{typetele, hato} for dexterous hand teleoperation, our system leverages motion-capture gloves to provide more precise and intuitive hand control. When combined with arm teleoperation, prior approaches often coupled VR joysticks (or similar localization devices) directly with the gloves~\cite{typetele, dexcap}, which we found to greatly restrict robotic arm rotation during teleoperation. In contrast, our decoupled design allows greater flexibility in arm motion. Furthermore, inspired by~\cite{rh20t}, we introduce a ``pause'' mechanism using a joystick button that lets the operator pause arm teleoperation at any time, reposition the joystick, and then resume control, as illustrated in Fig.~\ref{fig:system}. This further improves the flexibility of our system, making it particularly suited for tasks that require substantial arm rotations, as demonstrated in \S\ref{sec:user-study}.

\subsection{Dexterous Hand State Quantization}\label{sec:method-quantize}

We obtain $N$ demonstrations during data collection, where each demonstration is a trajectory $\{(o_i, s_i^{(a)}, s_i^{(h)})\}_i$, and $o_i$, $s_i^{(a)}$, $s_i^{(h)}$ denote the observation, arm state, and hand state at time step $i$, respectively. Previous methods~\cite{vqbet, vqvla, vqace} typically quantize concatenated arm-hand action chunks $\{(s_{i+k}^{(a)}, s_{i+k}^{(h)})\}_{k=1}^{C}$ with chunk size $C$. We argue, however, that this design is inappropriate for two reasons.

\textbf{First, we should quantize hand actions only, rather than concatenated arm-hand actions (Fig.~\ref{fig:baseline}C)} As discussed in \S\ref{sec:intro}, arm and hand actions serve fundamentally different purposes: the arm primarily manages spatial localization, while the hand governs interaction once the target region is reached. For example, when opening a jar, the arm must first move the hand to the correct position on the lid before the fingers perform the hooking action. Precise arm control is therefore essential for enabling proper finger interaction, whereas small inaccuracies in hand control are often tolerable. This makes hand actions a more natural target for quantization than concatenated arm-hand actions.

\textbf{Second, we should quantize single-step hand actions, \textit{i.e.}, hand states, rather than hand action chunks (Fig.~\ref{fig:baseline}D)} Hand action chunks capture temporal motion patterns of the dexterous hand, so quantizing them directly encodes these patterns into the codebook~\cite{vqace}, leading to rapid codebook expansion compared to state-level quantization. Importantly, when hand action chunks are quantized, they must be classified from a discrete chunk codebook, whereas arm action chunks are typically generated via the diffusion process~\cite{dp,rise,rise2,pi0}. This mismatch in action generation methods disentangles the two processes and can severely disrupt arm-hand coordination. For example, when opening a jar, the fingers' hooking motion should only occur once the arm has correctly positioned the hand on the lid. If the hand motion is triggered too early due to the separation of arm and hand action chunk generation, the fingers may miss the lid or collide with the jar, causing manipulation failures.

To avoid these issues, we adopt single-step hand action quantization, \textit{i.e.}, hand state quantization. Concretely, we extract hand states ${s^{(h)}}$ from the demonstration dataset $\mathcal{D}$ and train a two-layer residual VQ-VAE~\cite{resvqvae} to discretize them, as shown in Fig.~\ref{fig:policy}\raisebox{1.4pt}{\resizebox{0.8\width}{0.8\height}{\textcircled{\raisebox{-0.9pt}{1}}}}. Each hand state $s^{(h)}$ is encoded into a latent $z_e$, quantized by nearest-neighbor lookup in hierarchical codebooks $\{z_q\}$, and then decoded to reconstruct $\hat{s}^{(h)}$. The model is optimized with the standard VQ-VAE loss,
\[
\mathcal{L} = \| s^{(h)} - \hat{s}^{(h)} \|_2^2 
+ \beta \| \text{sg}[z_e] - z_q \|_2^2 
+ \gamma \| z_e - \text{sg}[z_q] \|_2^2 ,
\]
where $\text{sg}[\cdot]$ denotes the stop-gradient operator, and $\beta$, $\gamma$ are weighting coefficients. The first term enforces reconstruction, while the latter two promote stable codebook usage.

\subsection{Continuous Relaxation of Discretized Hand State}\label{sec:method-continuous}

After quantizing hand states, we merge the multi-layer codebooks into a unified codebook with $K$ discrete hand state codes. A natural approach is to formulate hand action prediction as a classification problem and use a classification head to predict future hand actions. However, as discussed in \S\ref{sec:method-quantize} and validated in our experiments, decoupling arm and hand action generation often leads to mismatched actions and degraded performance. Hence, we consider \textit{integrating the discrete hand states into arm action chunk diffusion}. We draw inspiration from the gripper: in practice, we usually only focus on whether it is open or closed, yet the visuomotor policy predicts a continuous value, as the transition from fully closed to fully open is inherently smooth and consistent. By analogy, if a consistent and continuous direction can be identified in the discrete hand state space, these states can likewise be predicted in a continuous manner and seamlessly integrated into the diffusion process.

To this end, we propose a continuous relaxation of the discretized hand states by re-indexing them in a continuous order. Instead of operating in the VQ-VAE latent space, we directly apply principal component analysis (PCA)~\cite{pca} to the raw 6-DoF hand states. Projecting onto the first principal component, which captures the largest variance, provides a one-dimensional representation that reflects the dominant trend of hand motion. The quantized hand states are then sequentially re-indexed along this axis, as shown in Fig.~\ref{fig:policy}\raisebox{1.4pt}{\resizebox{0.8\width}{0.8\height}{\textcircled{\raisebox{-0.9pt}{2}}}}.

This design ensures that neighboring states in the re-indexed sequence correspond to similar hand configurations in the original hand state space, yielding a coherent ordering of gestures. By contrast, applying PCA on high-dimensional VQ-VAE features does not guarantee that adjacent indices represent semantically consistent hand poses. Our approach therefore provides a more interpretable continuous relaxation of the discretized hand states.

\subsection{Visuomotor Policy Learning}\label{sec:method-policy}

After re-indexing the discretized hand states, we relabel each hand action $a^{(h)}$ in the dataset with its corresponding quantized, ordered index $z^{(h)}$. The demonstration trajectory can thus be represented as $\{(o_i, (s_i^{(a)}, z_i^{(h)}))\}_i$, as shown in Fig.~\ref{fig:policy}\raisebox{1.4pt}{\resizebox{0.8\width}{0.8\height}{\textcircled{\raisebox{-0.9pt}{3}}}}. We then train a base visuomotor policy by using the observation $o_i$ as input and a chunk of concatenated future arm and re-indexed hand actions $\{(s_{i+k}^{(a)}, z_{i+k}^{(h)})\}_{k=1}^C$ as output, as shown in Fig.~\ref{fig:policy}\raisebox{1.4pt}{\resizebox{0.8\width}{0.8\height}{\textcircled{\raisebox{-0.9pt}{4}}}}. During inference, the predicted continuous hand action $\hat{z}^{(h)}$ is mapped to its nearest quantized code $\text{idx}=[\hat{z}^{(h)}]$, from which the corresponding hand state $s^{(h)}_{\text{idx}}$ is retrieved for execution.

We adopt RISE~\cite{rise} as the base visuomotor policy for its strong spatial generalization. Point clouds from the two cameras are first combined using the calibrated extrinsics, and cropped to the workspace region, then fed into the RISE policy to predict future arm-hand action chunks.

\begin{figure}[t]
    \centering
    \includegraphics[width=\linewidth]{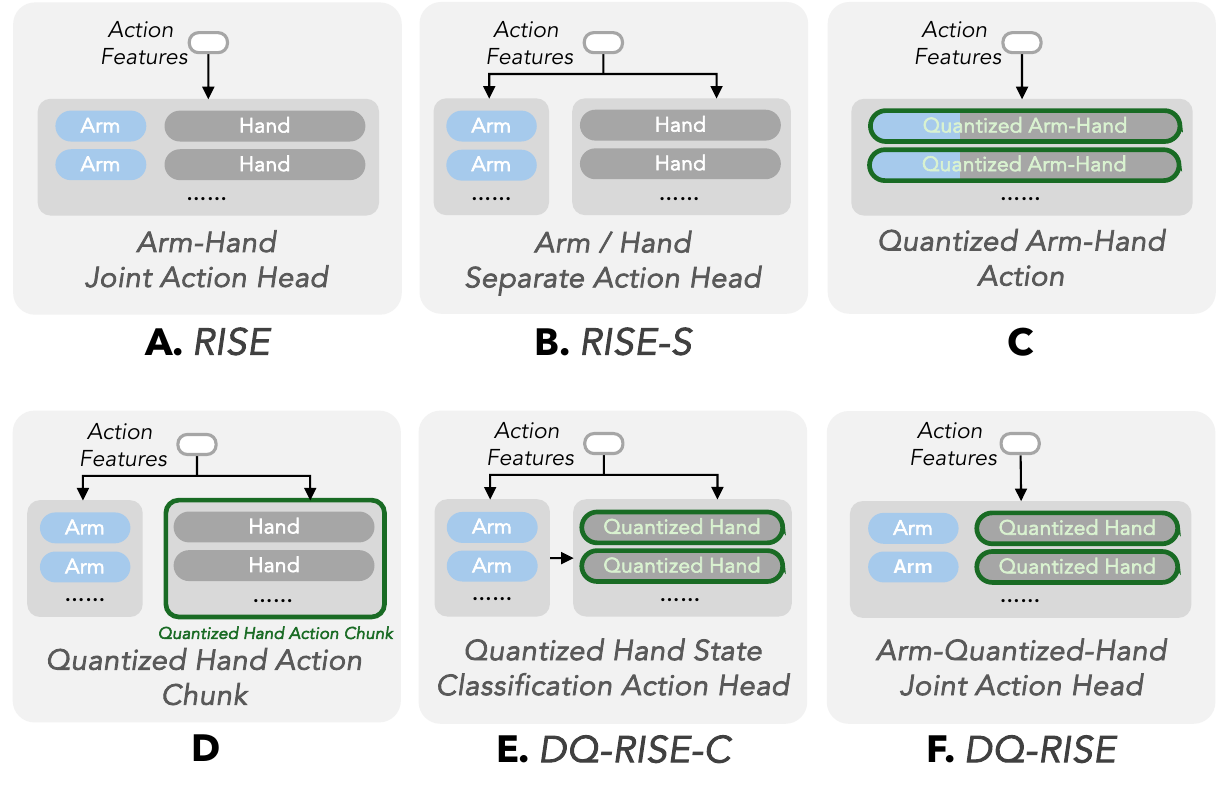}
    \caption{\textbf{Different Action Prediction Frameworks.} We select \textit{RISE}, \textit{RISE-S}, \textit{DQ-RISE-C} as baselines and compare with our \textit{DQ-RISE}.}
    \label{fig:baseline}\vspace{-0.4cm}
\end{figure}

\section{Experiments}

\begin{figure*}
    \centering
    \includegraphics[width=\linewidth]{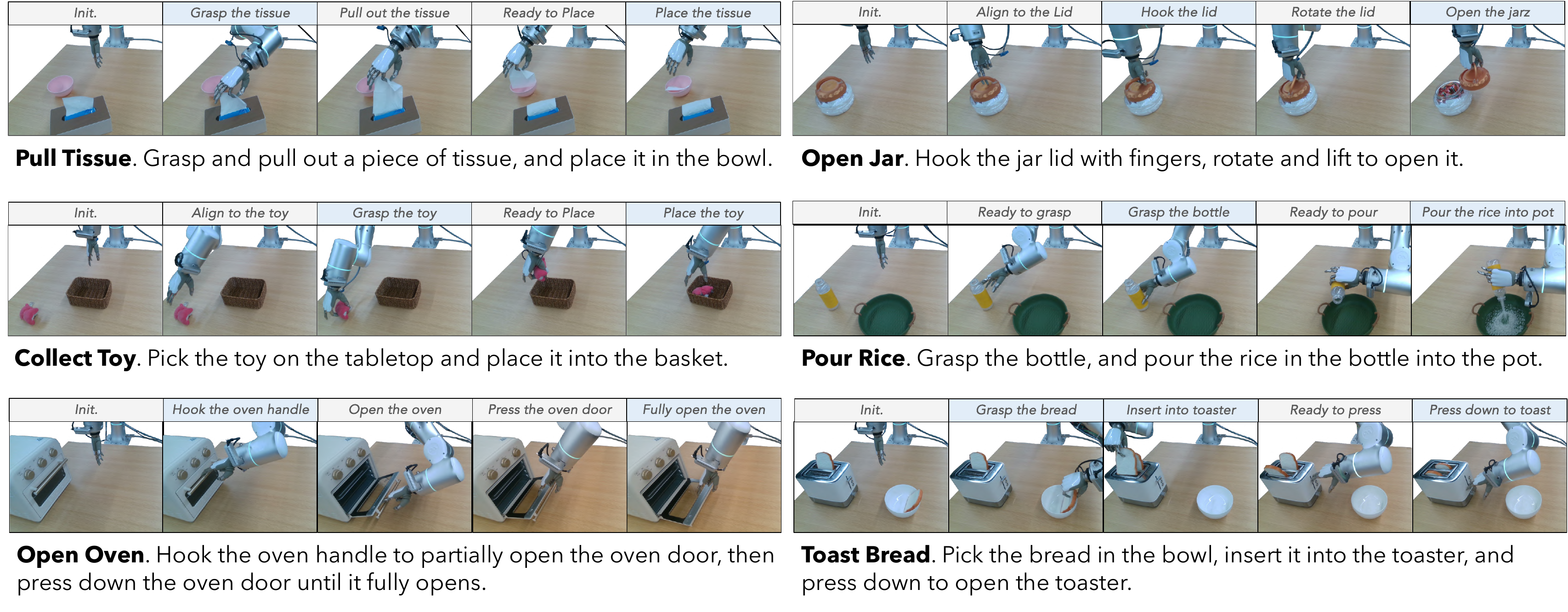}\vspace{-0.1cm}
    \caption{\textbf{Task Descriptions.} We evaluate six tasks covering pick-and-place (\textbf{\textit{Pull Tissue}}, \textbf{\textit{Collect Toy}}), articulated object manipulation (\textbf{\textit{Open Jar}}, \textbf{\textit{Open Oven}}), tasks requiring large rotations (\textbf{\textit{Open Jar}}, \textbf{\textit{Pour Rice}}), and a long-horizon task (\textbf{\textit{Toast Bread}}). Each task is illustrated with several phases, with the stages used for success rate evaluations highlighted in blue.}
    \label{fig:tasks}
\end{figure*}

\begin{table*}
    \centering
    \begin{tabular}{l rr rr rr rr rr rrr r}
         \toprule
         \multirow{2}{*}{\textbf{Policy}} & \multicolumn{2}{c}{\textbf{\textit{Pull Tissue}}} & \multicolumn{2}{c}{\textbf{\textit{Open Jar}}} & \multicolumn{2}{c}{\textbf{\textit{Collect Toy}}} & \multicolumn{2}{c}{\textbf{\textit{Pour Rice}}} & \multicolumn{2}{c}{\textbf{\textit{Open Oven}}} & \multicolumn{3}{c}{\textbf{\textit{Toast Bread}}} & \multicolumn{1}{c}{\multirow{2}{*}{\textbf{Avg.}}}\\ \cmidrule(lr){2-3}\cmidrule(lr){4-5}\cmidrule(lr){6-7}\cmidrule(lr){8-9}\cmidrule(lr){10-11}\cmidrule(lr){12-14}
         & \multicolumn{1}{c}{\textit{Grasp}} & \multicolumn{1}{c}{\textit{Place}} & \multicolumn{1}{c}{\textit{Hook}} & \multicolumn{1}{c}{\textit{Open}} & \multicolumn{1}{c}{\textit{Grasp}} & \multicolumn{1}{c}{\textit{Place}} & \multicolumn{1}{c}{\textit{Grasp}} & \multicolumn{1}{c}{\textit{Pour}} & \multicolumn{1}{c}{\textit{Hook}} &  \multicolumn{1}{c}{\textit{Press}} & \multicolumn{1}{c}{\textit{Grasp}} & \multicolumn{1}{c}{\textit{Insert}} & \multicolumn{1}{c}{\textit{Press}} \\ \midrule
         \textit{RISE}~\cite{rise} & 75\% & 45\% & 80\% & 55\% & 60\% & 60\% & 90\% & 80\% & \textbf{100}\% & 90\% & 80\% & 20\% & 0\% & 55.00\%\\
         \textit{RISE-S} & 75\% & 55\% & 60\% & 45\% & 75\% & 70\% & 95\% & 85\% & 95\% & 95\% & 75\% & 25\% & 20\% & 61.67\% \\
         \textit{DQ-RISE-C} & 15\% & 10\% & 0\% & 0\% & 0\% & 0\% & 0\% & 0\% & 20\% & 5\% & 0\% & 0\% & 0\% & 2.50\%\\
         \textit{\textbf{DQ-RISE} (ours)} & \textbf{95}\% & \textbf{85}\% & \textbf{95}\% & \textbf{90}\% & \textbf{95}\% & \textbf{80}\% & \textbf{100}\% & \textbf{100}\% & \textbf{100}\% & \textbf{100}\% & \textbf{100}\% & \textbf{65}\% & \textbf{60}\% & \textbf{85.83}\% \\
         \bottomrule
    \end{tabular}
    \caption{\textbf{Evaluation Results.} We report the success rates of every policy in certain task phases (Fig.~\ref{fig:tasks}). \textbf{\textit{DQ-RISE}} outperforms other action prediction variants and can complete various dexterous manipulation tasks effectively, even on the challenging \textbf{\textit{Toast Bread}} task.}
    \label{tab:result}
    \vspace{-0.4cm}
\end{table*}

In the experiments section, we aim to address the following research questions: \textbf{(Q1)} Can our \textbf{\textit{DQ-RISE}} policy handle diverse dexterous manipulation tasks? \textbf{(Q2)} Which action prediction scheme yields the best performance in visuomotor policies for dexterous manipulation? \textbf{(Q3)} Is the continuous relaxation process  of the \textbf{\textit{DQ-RISE}} policy essential for effective learning? \textbf{(Q4)} How does manual hand-state quantization during data collection differ from our automatic quantization in policy training? \textbf{(Q5)} Does our VR-glove hybrid teleoperation system provide a more intuitive single-arm teleoperation interface for dexterous hands?

\subsection{Setup} 

\textbf{Tasks.} We design 6 tasks to evaluate the policies on dexterous manipulation in the real world, including pick-and-place operations, articulated object manipulation, tasks with significant rotations, and long-horizon tasks. Please refer to Fig.~\ref{fig:tasks} for detailed descriptions of each task.

\textbf{Baselines.} We compare our \textbf{\textit{DQ-RISE}} policy against three baselines for integrating dexterous hand action prediction: (1) the base visuomotor policy (\textit{RISE}, Fig.~\ref{fig:baseline}A), which predicts concatenated arm-hand action chunks; (2) the base visuomotor policy with separate diffusions (\textit{RISE-S}, Fig.~\ref{fig:baseline}B), which uses two diffusion heads for arm and hand action generation; and (3) the base visuomotor policy with quantized hand action classification (\textit{DQ-RISE-C}, Fig.~\ref{fig:baseline}E), which diffuses arm actions first and then classifies quantized hand actions using action features and the predicted arm action. We omit other baselines (Fig.~\ref{fig:baseline}C and Fig.~\ref{fig:baseline}D) according to the previous discussions in \S\ref{sec:method-quantize}.

\textbf{Implementations.} We train a two-layer residual VQ-VAE~\cite{resvqvae} with a codebook size of 4 for each layer, resulting in $K=16$ quantized hand states per task. We set the both the commitment cost and codebook usage weight $\beta=\gamma=1.67$. The residual VQ-VAE is optimized using Adam~\cite{adam} with a learning rate of $3\times10^{-4}$, a batch size of 256 for 1500 epochs. Other policy hyperparameters follow RISE~\cite{rise}.

\textbf{Protocols.} We use our VR-glove hybrid dexterous teleoperation system (\S\ref{sec:data-collection}) to collect 50 teleoperated demonstrations per task, and use these demonstrations to train the policies. We deploy the policies on a workstation with an NVIDIA RTX 3090 GPU. Following~\cite{umi, cage, rise2}, object positions are randomized within the workspace before each task. Each policy is evaluated over 20 trials per task, and we report the overall success rate as the primary task completion metric, along with success rates broken down by task phases.

\subsection{Results}\label{sec:exp-result}

\textbf{\textit{DQ-RISE} policy is able to effectively handle a wide range of dexterous manipulation tasks (Q1).} It achieves the highest success rates across all six evaluated tasks, with an average success rate of 85.83\%. Beyond basic pick-and-place operations and articulated object manipulation such as opening an oven or a jar, \textbf{\textit{DQ-RISE}} can also perform more complex long-horizon tasks like \textbf{\textit{Toast Bread}}. Many of these tasks require the hand to adaptively switch between distinct poses at different stages. For example, in the \textbf{\textit{Toast Bread}} task, the robot first uses its thumb and index finger to grasp the bread, and after placing it into the toaster, employs its index and middle fingers to press the button. Importantly, our quantized hand states prove sufficient for such tasks, reinforcing our design motivation: \textit{the arm primarily handles localization, while the hand only needs to memorize certain action patterns}. Our quantization dramatically reduces the size of the hand state space, enabling the policy to concentrate on the more challenging problem of precise arm localization during manipulation.

\textbf{Our action prediction scheme with joint arm and quantized hand action prediction, achieves the best performance among all alternatives (Q2).} This result further demonstrates that quantizing hand states facilitates effective arm action learning and arm-hand coordination, while still preserving fine-grained hand actions to a large extent. In contrast, the vanilla approach of directly concatenating arm and hand action spaces (\textit{RISE}) struggles with fine-grained localization. For example, in the \textbf{\textit{Pull Tissue}} and \textbf{\textit{Collect Toys}} tasks, accurate localization is essential not only for grasping the tissue or toy but also for placing them precisely into the bowl or basket. These difficulties verify our hypothesis from \S\ref{sec:intro} that the high-DoF hand state tends to dominate the action space, making learning less effective. Naively separating the arm and hand action predictions (\textit{RISE-S}) alleviates this issue and improves performance on most tasks, but fails on the \textbf{\textit{Open Jar}} task, where tight arm-hand coordination is crucial to hook and rotate the lid, as illustrated in Fig.~\ref{fig:teaser}. 

\textbf{Combining classification with diffusion-based action generation on the same conditioning feature can introduce inconsistent gradient flows during training, ultimately degrading policy performance (Q2).} In our experiments, \textit{DQ-RISE-C} rarely succeeds in completing tasks. We observe that while the policy often predicts arm actions quite well, failures in hand state classification --- such as prematurely changing the hand pose --- distort subsequent observations, push the arm trajectory out of distribution, and ultimately cause rollouts to collapse. One possible explanation is that, since quantized hand actions are classified conditioned on the predicted arm action, distribution shift in arm predictions during inference could harm classification. However, an ablation study on whether to include this arm-conditioning route (Fig. \ref{fig:ablation}A) shows a negligible effect, ruling out this factor. We thus attribute the failure primarily to inconsistent gradient flows between the arm diffusion head and the hand classification head, which hinder effective joint optimization and result in suboptimal learning. Similar interference between classification and regression objectives has been documented in multi-task learning~\cite{mt1,mt2}, and comparable observations have been made in other visuomotor policies~\cite{foar}. These findings further validate our design choice to apply continuous relaxation to quantized hand states, avoiding a classification objective and ensuring consistent gradient propagation during training.

\subsection{Ablations}\label{sec:ablation}

\begin{figure}[t]
    \centering
    \includegraphics[width=\linewidth]{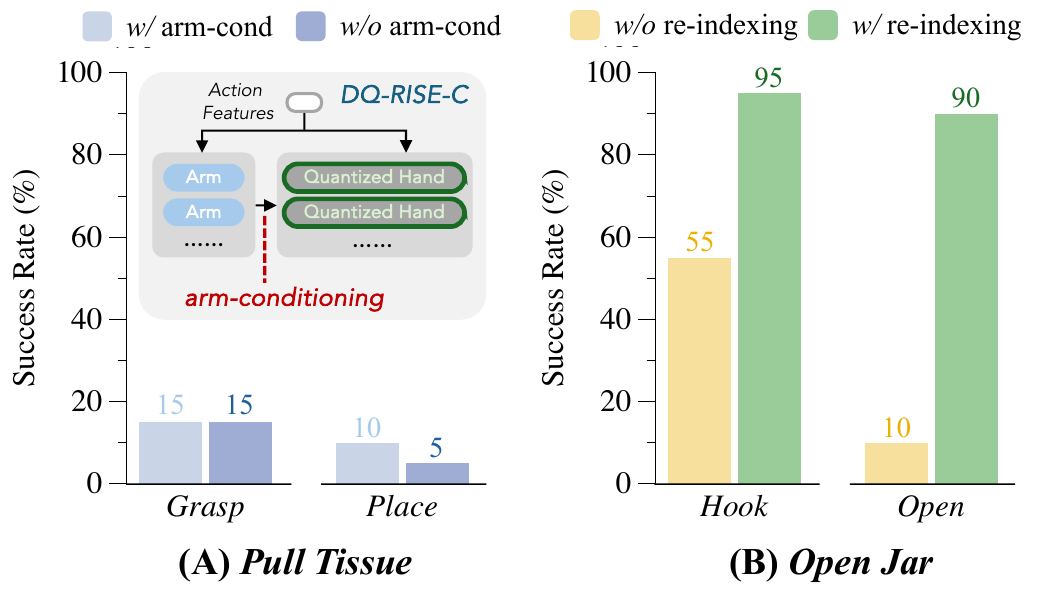}\vspace{-0.15cm}
    \caption{\textbf{(A) Ablation of Arm Conditioning in \textit{DQ-RISE-C}.} The policies perform similarly regardless of whether arm conditioning is applied during hand code classification. \textbf{(B) Ablation of Continuous Relaxation in \textit{DQ-RISE}.} Perform continuous relaxation (\textit{i.e.}, re-indexing) enhances the interpretability of the predicted hand codes, and significantly boosts performance.}
    \label{fig:ablation}\vspace{-0.35cm}
\end{figure}

\textbf{The continuous relaxation process is essential for effective policy learning (Q3).} We select the \textbf{\textit{Open Jar}} task as an example to ablate the function of the continuous relaxation process. As shown in Fig.~\ref{fig:ablation}B, removing re-indexing leads to a substantial drop in policy performance, whereas our policy achieves a much higher success rate. Without continuous ordering, neighboring code indices may correspond to very different hand states, making policy learning difficult and unstable. Moreover, non-continuous codes reduce the policy's tolerance to prediction errors: even a small mistake could map to a drastically different hand configuration. In contrast, continuous code indices ensure that nearby predictions correspond to similar hand states, improving robustness and allowing the policy to coordinate arm and hand actions more reliably. This confirms that continuous relaxation is a key component in \textbf{\textit{DQ-RISE}}.

\begin{figure}[t]
    \centering
    \includegraphics[width=0.98\linewidth]{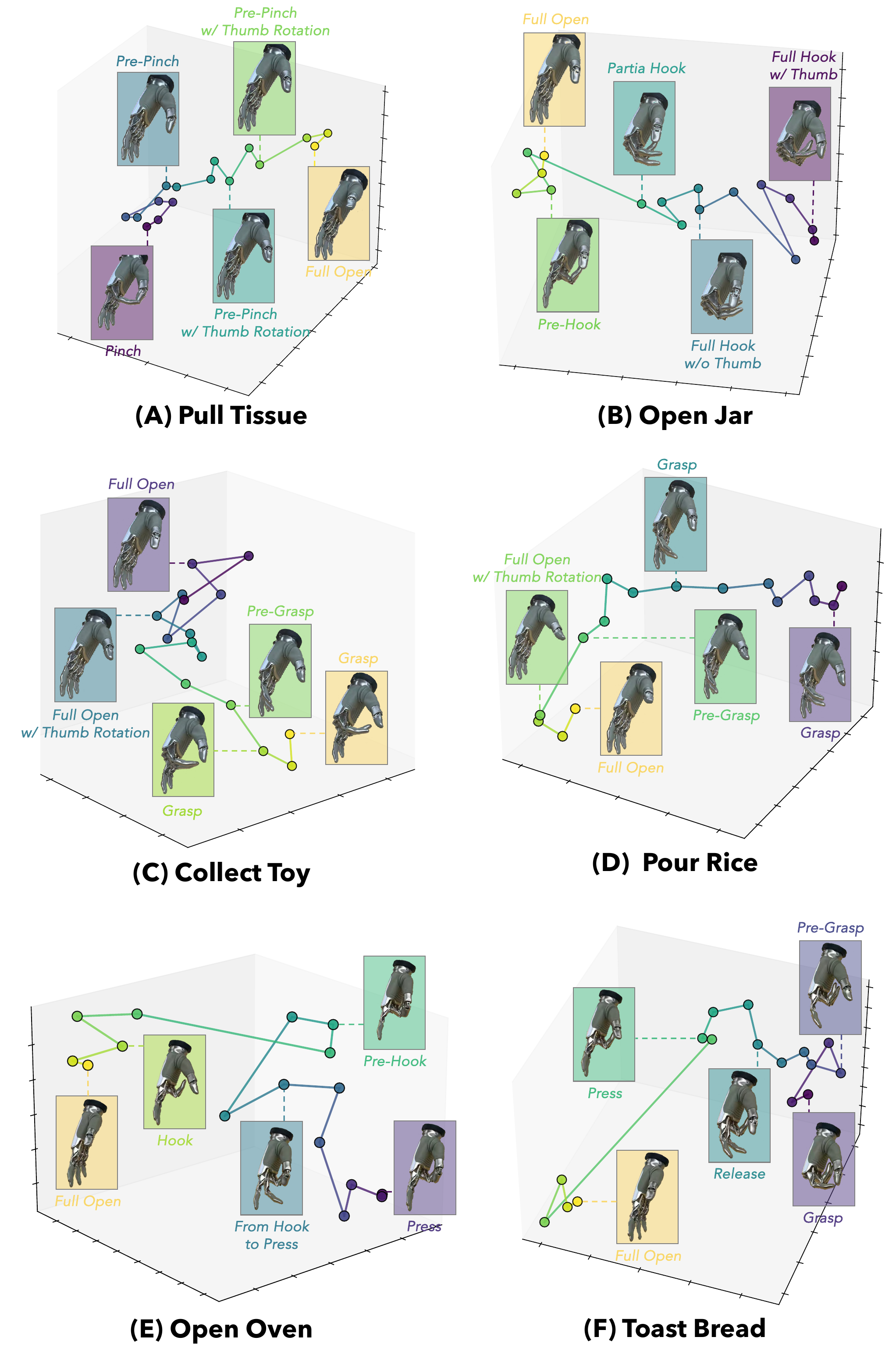}
    \caption{\textbf{Quantized Hand State after Re-Indexing.} Hand states are projected into 3D points via UMAP~\cite{umap}, with selected points annotated by their corresponding hand poses for reference. Re-indexing in the continuous relaxation process makes code transitions continuous and interpretable in the hand states, supporting further joint arm action and quantized hand action diffusing.}
    \label{fig:quantize}
    \vspace{-0.45cm}
\end{figure}

\textbf{The continuous relaxation process makes quantized hand state codes interpretable and allows their indices to be predicted continuously (Q3).} As illustrated in Fig.~\ref{fig:quantize}, neighboring code indices correspond to smoothly varying hand poses, and interpolation between indices yields meaningful intermediate hand configurations. This continuous, structured representation enables the policy to predict arm and hand actions jointly, without needing separate treatments for arm prediction and hand prediction. 

\subsection{User Study}\label{sec:user-study}

\begin{table}[b]
    \centering
    \setlength\tabcolsep{4.5pt}
    \begin{tabular}{lccc}
        \toprule
        \multirow{2}{*}{\textbf{Teleoperation System}} & \multicolumn{2}{c}{\textit{\textbf{Open Jar}}} & \multirow{2}{*}{\begin{tabular}{c}\textbf{Avg.}\\ \textbf{Rank} $\downarrow$ \end{tabular}} \\ \cmidrule(lr){2-3}
         & \textit{Success Rate} $\uparrow$ & \textit{Time} (s) $\downarrow$ & \\
         \midrule
         Coupled arm-hand control & 5 / 6 & 25.17 & 3.83 \\\midrule
         Ours \textit{w/} discretized gesture & \textbf{6} / 6 & 20.50 & 2.83 \\
         Ours \textit{w/o} pausing & \textbf{6} / 6 & 16.67 & 2.25 \\
         Ours & \textbf{6} / 6 & \textbf{13.83} & \textbf{1.08}\\
         \bottomrule
    \end{tabular}
    \caption{\textbf{User Study Results.} Our arm-hand decoupled teleoperation system with a pausing mechanism is both intuitive and convenient to control the arm and the dexterous hand, improving success rates and reducing completion time during teleoperation. }
    \label{tab:user_study}
    \vspace{-0.354cm}
\end{table}

We conduct a user study to evaluate our VR-glove hybrid dexterous teleoperation system against three alternatives: (1) \textit{a coupled arm-hand control system}, where the VR joystick is attached to the user's wrist and the glove controls the robotic hand; (2) \textit{a discretized gesture system}, in which users control the hand via discrete gestures (\textit{e.g.}, keyboard inputs) while using the same VR device for arm motion; and (3) \textit{a variant without a pausing mechanism}, which does not allow pausing the arm during teleoperation. Six participants with varying teleoperation experience has 3 minutes to familiarize themselves with each system before attempting the \textbf{\textit{Open Jar}} task in randomized order. Afterward, they are asked to rank the systems by intuitiveness and convenience.

\textbf{Our VR-glove hybrid teleoperation system provides a more intuitive single-arm interface for dexterous hands (Q5).} As shown in Tab.~\ref{tab:user_study}, it achieves the highest success rate (6/6), shortest completion time (13.83s), and best average rank (1.08). The coupled arm-hand system performs the worst due to the severe rotations required for the \textbf{\textit{Open Jar}} task, making the teleoperation control difficult. The discretized gesture and no-pausing variants improve over it, but still fall short in intuitiveness and convenience, respectively. These results demonstrate that integrating continuous glove-based hand control with VR-based arm control and a pausing mechanism enhances efficiency, reduces cognitive load, and improves coordination, confirming our system's superiority for single-arm dexterous manipulation.

\section{Conclusion}

In this work, we explore visuomotor policy learning on dexterous manipulations from the action prediction perspective. Our study highlights that successful manipulation depends not only on fine-grained hand motions but also on precise arm localization and coherent arm-hand coordination. Existing approaches, which either combine or fully separate arm and hand action spaces, often suffer from imbalanced learning or degraded coordination. To address these challenges, we propose \textit{\textbf{DQ-RISE}}, which quantizes dexterous hand states into a compact set of task-relevant patterns and jointly diffuses them with arm actions, allowing the policy to focus on accurate arm localization while retaining dexterous hand capabilities. We also introduce a hybrid dexterous teleoperation system to support intuitive and convenient demonstration collection. Experiments across diverse dexterous tasks show that \textit{\textbf{DQ-RISE}} consistently achieves the best overall performance, validating the effectiveness of our structured action prediction framework.

Looking ahead, we see two promising directions for future research. First, extending \textbf{\textit{DQ-RISE}} to multi-task learning presents challenges, as hand state quantization may introduce diverse hand states as codes when task distributions vary widely, potentially requiring adaptive or hierarchical quantization schemes. Second, the continuous relaxation that underpins joint arm-hand diffusion could become more difficult to stabilize in multi-task or long-horizon settings, motivating the exploration of improved relaxation techniques or hybrid discrete-continuous formulations. Addressing these challenges would further advance the scalability and generalization of visuomotor policies, paving the way for robust and versatile dexterous manipulation in real-world environments.

\section*{Acknowledgement}

This work was supported in part by the National Natural Science Foundation of China (No. 62595774), the Shanghai Committee of Science and Technology, China (Grant No. 24511103200), Shanghai Artificial Intelligence Laboratory, XPLORER PRIZE grants.

\textit{\textbf{Contributions}: H. Fang initiated the project with Y. Feng. H. Fang, Y. Feng, and Y. He designed the policy architecture and evaluation tasks. Y. Feng set up the teleoperation and data collection system. Y. Feng and Y. He trained the network and conducted the evaluation. J. Chen and C. Wang helped with the user study. H. Fang devised and mentored this project. H. Fang, Y. Feng, Y. He, and J. Chen wrote the paper. Z. He helped during the idea development. R. Luo and C. Lu supervised the project and provided hardware support.}
 
\printbibliography

@inproceedings{bidex,
  author       = {Kenneth Shaw and
                  Yulong Li and
                  Jiahui Yang and
                  Mohan Kumar Srirama and
                  Ray Liu and
                  Haoyu Xiong and
                  Russell Mendonca and
                  Deepak Pathak},
  title        = {Bimanual Dexterity for Complex Tasks},
  booktitle    = {CoRL},
  pages        = {5166--5183},
  year         = {2024}
}

@article{vitacformer,
  title={ViTacFormer: Learning Cross-Modal Representation for Visuo-Tactile Dexterous Manipulation},
  author={Heng, Liang and Geng, Haoran and Zhang, Kaifeng and Abbeel, Pieter and Malik, Jitendra},
  journal={arXiv preprint arXiv:2506.15953},
  year={2025}
}

@inproceedings{maniptrans,
  author       = {Kailin Li and
                  Puhao Li and
                  Tengyu Liu and
                  Yuyang Li and
                  Siyuan Huang},
  title        = {ManipTrans: Efficient Dexterous Bimanual Manipulation Transfer via
                  Residual Learning},
  booktitle    = {{CVPR}},
  pages        = {6991--7003},
  year         = {2025}
}

@inproceedings{dexmv,
  author       = {Yuzhe Qin and
                  Yueh{-}Hua Wu and
                  Shaowei Liu and
                  Hanwen Jiang and
                  Ruihan Yang and
                  Yang Fu and
                  Xiaolong Wang},
  title        = {DexMV: Imitation Learning for Dexterous Manipulation from Human Videos},
  booktitle    = {ECCV},
  pages        = {570--587},
  year         = {2022}
}

@inproceedings{open-television,
  author       = {Xuxin Cheng and
                  Jialong Li and
                  Shiqi Yang and
                  Ge Yang and
                  Xiaolong Wang},
  title        = {Open-TeleVision: Teleoperation with Immersive Active Visual Feedback},
  booktitle    = {CoRL},
  pages        = {2729--2749},
  year         = {2024}
}

@inproceedings{hato,
  title={Learning Visuotactile Skills With Two Multifingered Hands},
  author={Lin, Toru and Zhang, Yu and Li, Qiyang and Qi, Haozhi and Yi, Brent and Levine, Sergey and Malik, Jitendra},
  booktitle={ICRA},
  pages={5637-5643},
  year={2025}
}

@inproceedings{dexcap,
  author       = {Chen Wang and
                  Haochen Shi and
                  Weizhuo Wang and
                  Ruohan Zhang and
                  Li Fei{-}Fei and
                  C. Karen Liu},
  title        = {DexCap: Scalable and Portable Mocap Data Collection System for Dexterous
                  Manipulation},
  booktitle    = {RSS},
  year         = {2024}
}

@inproceedings{leap_hand,
  author       = {Kenneth Shaw and
                  Ananye Agarwal and
                  Deepak Pathak},
  title        = {{LEAP} Hand: Low-Cost, Efficient, and Anthropomorphic Hand for Robot Learning},
  booktitle    = {RSS},
  year         = {2023}
}

@inproceedings{tilde,
  author       = {Zilin Si and
                  Kevin Lee Zhang and
                  Zeynep Temel and
                  Oliver Kroemer},
  title        = {Tilde: Teleoperation for Dexterous In-Hand Manipulation Learning with a DeltaHand},
  booktitle    = {RSS},
  year         = {2024}
}

@inproceedings{eyesight_hand,
  author       = {Branden Romero and
                  Haoshu Fang and
                  Pulkit Agrawal and
                  Edward H. Adelson},
  title        = {EyeSight Hand: Design of a Fully-Actuated Dexterous Robot Hand with
                  Integrated Vision-Based Tactile Sensors and Compliant Actuation},
  booktitle    = {{IROS}},
  pages        = {1853--1860},
  year         = {2024}
}

@inproceedings{dex-survey,
  author       = {Allison M. Okamura and
                  Niels Smaby and
                  Mark R. Cutkosky},
  title        = {An Overview of Dexterous Manipulation},
  booktitle    = {ICRA},
  pages        = {255--262},
  year         = {2000}
}

@inproceedings{dexumi,
  author       = {Xu, Mengda and Zhang, Han and Hou, Yifan and Xu, Zhenjia and Fan, Linxi and Veloso, Manuela and Song, Shuran},
  title        = {DexUMI: Using Human Hand as the Universal Manipulation Interface for Dexterous Manipulation},
  booktitle    = {CoRL},
  year         = {2025}
}

@article{dexmachina,
  title={DexMachina: Functional Retargeting for Bimanual Dexterous Manipulation},
  author={Mandi, Zhao and Hou, Yifan and Fox, Dieter and Narang, Yashraj and Mandlekar, Ajay and Song, Shuran},
  journal={arXiv preprint arXiv:2505.24853},
  year={2025}
}

@article{dexos,
  title={DEXOP: A Device for Robotic Transfer of Dexterous Human Manipulation},
  author={Fang, Hao-Shu and Romero, Branden and Xie, Yichen and Hu, Arthur and Huang, Bo-Ruei and Alvarez, Juan and Kim, Matthew and Margolis, Gabriel and Anbarasu, Kavya and Tomizuka, Masayoshi and others},
  journal={arXiv preprint arXiv:2509.04441},
  year={2025}
}

@inproceedings{geort,
  title={Geometric Retargeting: A Principled, Ultrafast Neural Hand Retargeting Algorithm},
  author={Yin, Zhao-Heng and Wang, Changhao and Pineda, Luis and Bodduluri, Krishna and Wu, Tingfan and Abbeel, Pieter and Mukadam, Mustafa},
  booktitle={IROS},
  year={2025}
}

@article{umap,
  title={UMAP: Uniform Manifold Approximation and Projection},
  author={McInnes, Leland and Healy, John and Saul, Nathaniel and Gro{\ss}berger, Lukas},
  journal={Journal of Open Source Software},
  volume={3},
  number={29},
  pages={861},
  year={2018}
}

@inproceedings{ace,
  author       = {Shiqi Yang and
                  Minghuan Liu and
                  Yuzhe Qin and
                  Runyu Ding and
                  Jialong Li and
                  Xuxin Cheng and
                  Ruihan Yang and
                  Sha Yi and
                  Xiaolong Wang},
  title        = {{ACE:} {A} Cross-platform and visual-Exoskeletons System for Low-Cost Dexterous Teleoperation},
  booktitle    = {CoRL},
  pages        = {4895--4911},
  year         = {2024}
}

@inproceedings{vqvae,
  author       = {A{\"{a}}ron van den Oord and
                  Oriol Vinyals and
                  Koray Kavukcuoglu},
  title        = {Neural Discrete Representation Learning},
  booktitle    = {NeurIPS},
  pages        = {6306--6315},
  year         = {2017}
}

@inproceedings{vqvae2,
  author       = {Ali Razavi and
                  A{\"{a}}ron van den Oord and
                  Oriol Vinyals},
  title        = {Generating Diverse High-Fidelity Images with {VQ-VAE-2}},
  booktitle    = {NeurIPS},
  pages        = {14837--14847},
  year         = {2019}
}

@article{dex-inhand,
  author       = {Marcin Andrychowicz and
                  Bowen Baker and
                  Maciek Chociej and
                  Rafal J{\'{o}}zefowicz and
                  Bob McGrew and
                  Jakub Pachocki and
                  Arthur Petron and
                  Matthias Plappert and
                  Glenn Powell and
                  Alex Ray and
                  Jonas Schneider and
                  Szymon Sidor and
                  Josh Tobin and
                  Peter Welinder and
                  Lilian Weng and
                  Wojciech Zaremba},
  title        = {Learning Dexterous In-Hand Manipulation},
  journal      = {Int. J. Robotics Res.},
  volume       = {39},
  number       = {1},
  year         = {2020}
}

@inproceedings{dexgraspnet,
  author       = {Ruicheng Wang and
                  Jialiang Zhang and
                  Jiayi Chen and
                  Yinzhen Xu and
                  Puhao Li and
                  Tengyu Liu and
                  He Wang},
  title        = {DexGraspNet: {A} Large-Scale Robotic Dexterous Grasp Dataset for General Objects Based on Simulation},
  booktitle    = {ICRA},
  pages        = {11359--11366},
  year         = {2023}
}

@inproceedings{unidexgrasp,
  author       = {Yinzhen Xu and
                  Weikang Wan and
                  Jialiang Zhang and
                  Haoran Liu and
                  Zikang Shan and
                  Hao Shen and
                  Ruicheng Wang and
                  Haoran Geng and
                  Yijia Weng and
                  Jiayi Chen and
                  Tengyu Liu and
                  Li Yi and
                  He Wang},
  title        = {UniDexGrasp: Universal Robotic Dexterous Grasping via Learning Diverse
                  Proposal Generation and Goal-Conditioned Policy},
  booktitle    = {{CVPR}},
  pages        = {4737--4746},
  year         = {2023}
}

@article{anydexgrasp,
  title={AnyDexGrasp: General Dexterous Grasping for Different Hands with Human-level Learning Efficiency},
  author={Fang, Hao-Shu and Yan, Hengxu and Tang, Zhenyu and Fang, Hongjie and Wang, Chenxi and Lu, Cewu},
  journal={arXiv preprint arXiv:2502.16420},
  year={2025}
}

@inproceedings{inhand-re,
  author       = {Tao Chen and
                  Jie Xu and
                  Pulkit Agrawal},
  title        = {A System for General In-Hand Object Re-Orientation},
  booktitle    = {CoRL},
  pages        = {297--307},
  year         = {2021}
}

@article{visual-dex,
  author       = {Tao Chen and
                  Megha Tippur and
                  Siyang Wu and
                  Vikash Kumar and
                  Edward H. Adelson and
                  Pulkit Agrawal},
  title        = {Visual Dexterity: In-hand Reorientation of Novel and Complex Object Shapes},
  journal      = {Sci. Robotics},
  volume       = {8},
  number       = {84},
  year         = {2023}
}

@article{bidexhands,
  author       = {Yuanpei Chen and
                  Yiran Geng and
                  Fangwei Zhong and
                  Jiaming Ji and
                  Jiechuang Jiang and
                  Zongqing Lu and
                  Hao Dong and
                  Yaodong Yang},
  title        = {Bi-DexHands: Towards Human-Level Bimanual Dexterous Manipulation},
  journal      = {{IEEE} Trans. Pattern Anal. Mach. Intell.},
  volume       = {46},
  number       = {5},
  pages        = {2804--2818},
  year         = {2024}
}

@article{dex-assembly,
  author       = {Dong{-}Hyuk Lee and
                  Myoung{-}Su Choi and
                  Hyeonjun Park and
                  Ga{-}Ram Jang and
                  Jae{-}Han Park and
                  Ji{-}Hun Bae},
  title        = {Peg-in-Hole Assembly With Dual-Arm Robot and Dexterous Robot Hands},
  journal      = {{IEEE} Robotics Autom. Lett.},
  volume       = {7},
  number       = {4},
  pages        = {8566--8573},
  year         = {2022}
}

@article{dex-tool,
  author       = {Fan Yang and
                  Wenrui Chen and
                  Haoran Lin and
                  Sijie Wu and
                  Xin Li and
                  Zhiyong Li and
                  Yaonan Wang},
  title        = {Task-Oriented Tool Manipulation With Robotic Dexterous Hands: {A}
                  Knowledge Graph Approach From Fingers to Functionality},
  journal      = {{IEEE} Trans. Cybern.},
  volume       = {55},
  number       = {1},
  pages        = {395--408},
  year         = {2025}
}

@inproceedings{rise,
    title     = {RISE: 3D Perception Makes Real-World Robot Imitation Simple and Effective},
    author    = {Wang, Chenxi and Fang, Hongjie and Fang, Hao-Shu and Lu, Cewu},
    booktitle = {IROS}, 
    year      = {2024},
    pages     = {2870-2877}
}

@inproceedings{dp,
	title={Diffusion Policy: Visuomotor Policy Learning via Action Diffusion},
	author={Chi, Cheng and Feng, Siyuan and Du, Yilun and Xu, Zhenjia and Cousineau, Eric and Burchfiel, Benjamin and Song, Shuran},
	booktitle={RSS},
	year={2023}
}

@inproceedings{act,
	title={Learning Fine-Grained Bimanual Manipulation with Low-Cost Hardware},
	author={Zhao, Tony Z and Kumar, Vikash and Levine, Sergey and Finn, Chelsea},
	booktitle={RSS},
	year={2023}
}

@inproceedings{rise2,
  title   = {AirExo-2: Scaling up Generalizable Robotic Imitation Learning with Low-Cost Exoskeletons},
  author  = {Hongjie Fang and Chenxi Wang and Yiming Wang and Jingjing Chen and Shangning Xia and Jun Lv and Zihao He and Xiyan Yi and Yunhan Guo and Xinyu Zhan and Lixin Yang and Weiming Wang and Cewu Lu and Hao-Shu Fang},
  booktitle = {CoRL},
  year    = {2025}
}

@inproceedings{pi0,
  title={$\pi_0$: A Vision-Language-Action Flow Model for General Robot Control},
  author={Black, Kevin and Brown, Noah and Driess, Danny and Esmail, Adnan and Equi, Michael and Finn, Chelsea and Fusai, Niccolo and Groom, Lachy and Hausman, Karol and Ichter, Brian and others},
  booktitle={RSS},
  year={2025}
}

@article{anygrasp,
  author       = {Hao-Shu Fang and
                  Chenxi Wang and
                  Hongjie Fang and
                  Minghao Gou and
                  Jirong Liu and
                  Hengxu Yan and
                  Wenhai Liu and
                  Yichen Xie and
                  Cewu Lu},
  title        = {AnyGrasp: Robust and Efficient Grasp Perception in Spatial and Temporal
                  Domains},
  journal      = {{IEEE} Trans. Robotics},
  volume       = {39},
  number       = {5},
  pages        = {3929--3945},
  year         = {2023}
}

@inproceedings{graspness,
  author       = {Chenxi Wang and
                  Haoshu Fang and
                  Minghao Gou and
                  Hongjie Fang and
                  Jin Gao and
                  Cewu Lu},
  title        = {Graspness Discovery in Clutters for Fast and Accurate Grasp Detection},
  booktitle    = {ICCV},
  pages        = {15944--15953},
  year         = {2021}
}

@inproceedings{typetele,
  title={TypeTele: Releasing Dexterity in Teleoperation by Dexterous Manipulation Types},
  author={Lin, Yuhao and Wei, Yi-Lin and Liao, Haoran and Lin, Mu and Xing, Chengyi and Li, Hao and Zhang, Dandan and Cutkosky, Mark and Zheng, Wei-Shi},
  booktitle={CoRL},
  year={2025}
}

@inproceedings{doglove,
  title={DOGlove: Dexterous Manipulation with a Low-Cost Open-Source Haptic Force Feedback Glove},
  author={Zhang, Han and Hu, Songbo and Yuan, Zhecheng and Xu, Huazhe},
  booktitle={RSS},
  year={2025}
}

@inproceedings{dart,
  title={DART: Dexterous Augmented Reality Teleoperation Platform for Large-Scale Robot Data Collection in Simulation},
  author={Park, Younghyo and Bhatia, Jagdeep Singh and Ankile, Lars and Agrawal, Pulkit},
  journal={ICRA},
  pages={13883-13889},
  year={2025}
}

@inproceedings{uva,
title={Unified Video Action Model},
author={Li, Shuang and Gao, Yihuai and Sadigh, Dorsa and Song, Shuran},
booktitle={RSS},
year={2025}
}

@inproceedings{kamijo2024learning,
  title={Learning Variable Compliance Control from a Few Demonstrations for Bimanual Robot with Haptic Feedback Teleoperation System},
  author={Kamijo, Tatsuya and Beltran-Hernandez, Cristian C and Hamaya, Masashi},
  booktitle={IROS},
  pages={12663--12670},
  year={2024}
}

@inproceedings{vqvla,
  title={VQ-VLA: Improving Vision-Language-Action Models via Scaling Vector-Quantized Action Tokenizers},
  author={Wang, Yating and Zhu, Haoyi and Liu, Mingyu and Yang, Jiange and Fang, Hao-Shu and He, Tong},
  booktitle={ICCV},
  year={2025}
}

@inproceedings{vae,
  author       = {Diederik P. Kingma and
                  Max Welling},
  title        = {Auto-Encoding Variational Bayes},
  booktitle    = {{ICLR}},
  year         = {2014}
}

@inproceedings{gan,
  author       = {Ian J. Goodfellow and
                  Jean Pouget{-}Abadie and
                  Mehdi Mirza and
                  Bing Xu and
                  David Warde{-}Farley and
                  Sherjil Ozair and
                  Aaron C. Courville and
                  Yoshua Bengio},
  title        = {Generative Adversarial Nets},
  booktitle    = {NeurIPS},
  pages        = {2672--2680},
  year         = {2014}
}

@inproceedings{ddpm,
  title={Denoising Diffusion Probabilistic Models},
  author={Ho, Jonathan and Jain, Ajay and Abbeel, Pieter},
  booktitle={NeurIPS},
  pages={6840--6851},
  year={2020}
}

@inproceedings{vqbet,
  author       = {Seungjae Lee and
                  Yibin Wang and
                  Haritheja Etukuru and
                  H. Jin Kim and
                  Nur Muhammad (Mahi) Shafiullah and
                  Lerrel Pinto},
  title        = {Behavior Generation with Latent Actions},
  booktitle    = {{ICML}},
  publisher    = {OpenReview.net},
  year         = {2024}
}

@inproceedings{moto,
  title={Moto: Latent Motion Token as the Bridging Language for Robot Manipulation},
  author={Chen, Yi and Ge, Yuying and Li, Yizhuo and Ge, Yixiao and Ding, Mingyu and Shan, Ying and Liu, Xihui},
  booktitle={ICCV},
  year={2025}
}

@article{gr2,
  title={GR-2: A Generative Video-Language-Action Model with Web-Scale Knowledge for Robot Manipulation},
  author={Cheang, Chi-Lam and Chen, Guangzeng and Jing, Ya and Kong, Tao and Li, Hang and Li, Yifeng and Liu, Yuxiao and Wu, Hongtao and Xu, Jiafeng and Yang, Yichu and others},
  journal={arXiv preprint arXiv:2410.06158},
  year={2024}
}

@inproceedings{upvla,
  author       = {Jianke Zhang and
                  Yanjiang Guo and
                  Yucheng Hu and
                  Xiaoyu Chen and
                  Xiang Zhu and
                  Jianyu Chen},
  title        = {{UP-VLA:} {A} Unified Understanding and Prediction Model for Embodied Agent},
  booktitle      = {ICML},
  year         = {2025}
}

@inproceedings{lapa,
  author       = {Seonghyeon Ye and
                  Joel Jang and
                  Byeongguk Jeon and
                  Se June Joo and
                  Jianwei Yang and
                  Baolin Peng and
                  Ajay Mandlekar and
                  Reuben Tan and
                  Yu{-}Wei Chao and
                  Bill Yuchen Lin and
                  Lars Liden and
                  Kimin Lee and
                  Jianfeng Gao and
                  Luke Zettlemoyer and
                  Dieter Fox and
                  Minjoon Seo},
  title        = {Latent Action Pretraining from Videos},
  booktitle    = {{ICLR}},
  year         = {2025}
}

@article{molmoact,
  title={MolmoAct: Action Reasoning Models that Can Reason in Space},
  author={Lee, Jason and Duan, Jiafei and Fang, Haoquan and Deng, Yuquan and Liu, Shuo and Li, Boyang and Fang, Bohan and Zhang, Jieyu and Wang, Yi Ru and Lee, Sangho and others},
  journal={arXiv preprint arXiv:2508.07917},
  year={2025}
}

@inproceedings{cage,
  title={CAGE: Causal Attention Enables Data-Efficient Generalizable Robotic Manipulation},
  author={Xia, Shangning and Fang, Hongjie and Lu, Cewu and Fang, Hao-Shu},
    booktitle = {ICRA},
  year={2025},
  pages={13242-13249}
}

@inproceedings{gello,
  author       = {Philipp Wu and
                  Yide Shentu and
                  Zhongke Yi and
                  Xingyu Lin and
                  Pieter Abbeel},
  title        = {{GELLO:} {A} General, Low-Cost, and Intuitive Teleoperation Framework
                  for Robot Manipulators},
  booktitle    = {{IROS}},
  pages        = {12156--12163},
  year         = {2024}
}

@inproceedings{adam,
  author       = {Diederik P. Kingma and
                  Jimmy Ba},
  title        = {Adam: {A} Method for Stochastic Optimization},
  booktitle    = {{ICLR}},
  year         = {2015}
}

@article{resvqvae,
  author       = {Neil Zeghidour and
                  Alejandro Luebs and
                  Ahmed Omran and
                  Jan Skoglund and
                  Marco Tagliasacchi},
  title        = {SoundStream: An End-to-End Neural Audio Codec},
  journal      = {{IEEE} {ACM} Trans. Audio Speech Lang. Process.},
  volume       = {30},
  pages        = {495--507},
  year         = {2022}
}

@inproceedings{anyteleop,
  title     = {AnyTeleop: A General Vision-Based Dexterous Robot Arm-Hand Teleoperation System},
  author    = {Qin, Yuzhe and Yang, Wei and Huang, Binghao and Van Wyk, Karl and Su, Hao and Wang, Xiaolong and Chao, Yu-Wei and Fox, Dieter},
  booktitle = {RSS},
  year      = {2023}
}

@inproceedings{CaggianoDK23,
  author       = {Vittorio Caggiano and
                  Sudeep Dasari and
                  Vikash Kumar},
  title        = {MyoDex: {A} Generalizable Prior for Dexterous Manipulation},
  booktitle    = {{ICML}},
  pages        = {3327--3346},
  year         = {2023}
}

@article{pca,
  title={Analysis of A Complex of Statistical Variables into Principal Components.},
  author={Hotelling, Harold},
  journal={Journal of Educational Psychology},
  volume={24},
  number={6},
  pages={417},
  year={1933}
}

@inproceedings{Zhu0RLK19,
  author       = {Henry Zhu and
                  Abhishek Gupta and
                  Aravind Rajeswaran and
                  Sergey Levine and
                  Vikash Kumar},
  title        = {Dexterous Manipulation with Deep Reinforcement Learning: Efficient,
                  General, and Low-Cost},
  booktitle    = {{ICRA}},
  pages        = {3651--3657},
  year         = {2019}
}

@inproceedings{NagabandiKLK19,
  author       = {Anusha Nagabandi and
                  Kurt Konolige and
                  Sergey Levine and
                  Vikash Kumar},
  title        = {Deep Dynamics Models for Learning Dexterous Manipulation},
  booktitle    = {CoRL},
  pages        = {1101--1112},
  year         = {2019}
}

@inproceedings{QinHY0022,
  author       = {Yuzhe Qin and
                  Binghao Huang and
                  Zhao{-}Heng Yin and
                  Hao Su and
                  Xiaolong Wang},
  title        = {DexPoint: Generalizable Point Cloud Reinforcement Learning for Sim-to-Real
                  Dexterous Manipulation},
  booktitle    = {CoRL},
  pages        = {594--605},
  year         = {2022}
}

@inproceedings{lin2025sim,
  title={Sim-to-real reinforcement learning for vision-based dexterous manipulation on humanoids},
  author={Lin, Toru and Sachdev, Kartik and Fan, Linxi and Malik, Jitendra and Zhu, Yuke},
  booktitle={CoRL},
  year={2025}
}

@article{vqace,
  title={VQ-ACE: Efficient Policy Search for Dexterous Robotic Manipulation via Action Chunking Embedding},
  author={Yang, Chenyu and Liconti, Davide and Katzschmann, Robert K},
  journal={arXiv preprint arXiv:2411.03556},
  year={2024}
}

@inproceedings{umi,
  author       = {Cheng Chi and
                  Zhenjia Xu and
                  Chuer Pan and
                  Eric Cousineau and
                  Benjamin Burchfiel and
                  Siyuan Feng and
                  Russ Tedrake and
                  Shuran Song},
  title        = {Universal Manipulation Interface: In-The-Wild Robot Teaching Without In-The-Wild Robots},
  booktitle    = {RSS},
  year         = {2024}
}

@inproceedings{vqgan,
  author       = {Patrick Esser and
                  Robin Rombach and
                  Bj{\"{o}}rn Ommer},
  title        = {Taming Transformers for High-Resolution Image Synthesis},
  booktitle    = {{CVPR}},
  pages        = {12873--12883},
  year         = {2021}
}

@inproceedings{airexo,
  author       = {Hongjie Fang and
                  Haoshu Fang and
                  Yiming Wang and
                  Jieji Ren and
                  Jingjing Chen and
                  Ruo Zhang and
                  Weiming Wang and
                  Cewu Lu},
  title        = {AirExo: Low-Cost Exoskeletons for Learning Whole-Arm Manipulation
                  in the Wild},
  booktitle    = {ICRA},
  pages        = {15031--15038},
  year         = {2024}
}

@inproceedings{rh20t,
  author       = {Hao-Shu Fang and
                  Hongjie Fang and
                  Zhenyu Tang and
                  Jirong Liu and
                  Chenxi Wang and
                  Junbo Wang and
                  Haoyi Zhu and
                  Cewu Lu},
  title        = {{RH20T:} {A} Comprehensive Robotic Dataset for Learning Diverse Skills in One-Shot},
  booktitle    = {ICRA},
  pages        = {653--660},
  year         = {2024},
}

@inproceedings{mt1,
  author       = {Ozan Sener and
                  Vladlen Koltun},
  title        = {Multi-Task Learning as Multi-Objective Optimization},
  booktitle    = {NeurIPS},
  pages        = {525--536},
  year         = {2018}
}

@inproceedings{mt2,
  author       = {Tianhe Yu and
                  Saurabh Kumar and
                  Abhishek Gupta and
                  Sergey Levine and
                  Karol Hausman and
                  Chelsea Finn},
  title        = {Gradient Surgery for Multi-Task Learning},
  booktitle    = {NeurIPS},
  year         = {2020}
}

@article{foar,
  author       = {Zihao He and
                  Hongjie Fang and
                  Jingjing Chen and
                  Haoshu Fang and
                  Cewu Lu},
  title        = {FoAR: Force-Aware Reactive Policy for Contact-Rich Robotic Manipulation},
  journal      = {{IEEE} Robotics Autom. Lett.},
  volume       = {10},
  number       = {6},
  pages        = {5625--5632},
  year         = {2025}
}

@inproceedings{clutterdexgrasp,
  author       = {Zeyuan Chen and
                  Qiyang Yan and
                  Yuanpei Chen and
                  Tianhao Wu and
                  Jiyao Zhang and
                  Zihan Ding and
                  Jinzhou Li and
                  Yaodong Yang and
                  Hao Dong},
  title        = {ClutterDexGrasp: {A} Sim-to-Real System for General Dexterous Grasping in Cluttered Scenes},
  booktitle      = {CoRL},
  year         = {2025}
}

\end{document}